\documentclass{article}

\usepackage[final]{neurips_2021_safe_rl}

\usepackage[utf8]{inputenc} 
\usepackage[T1]{fontenc}    
\usepackage{hyperref}       
\usepackage{url}          
\usepackage{booktabs}      
\usepackage{nicefrac}      
\usepackage{microtype}      
\usepackage{graphicx}
\usepackage{amsmath}
\usepackage{amsthm}
\usepackage{amsfonts}
\usepackage{amssymb}
\usepackage{mathtools}
\usepackage{subfigure}

\usepackage{caption}
\usepackage{graphicx}

\usepackage{colortbl}
\usepackage{xcolor}
\usepackage{enumitem}
\usepackage{bm}
\usepackage{lineno}
\usepackage{textcomp}
\usepackage{gensymb}
\usepackage{flushend}
\usepackage{multirow}
\usepackage{times}
\usepackage{natbib}
\usepackage[font=it]{caption}
\usepackage{doi}
\usepackage{placeins}
\usepackage{psfrag}
\usepackage{siunitx}
\usepackage{eurosym, units}
\usepackage{xtab}
\usepackage{verbatim}
\usepackage{wrapfig}
\usepackage{pgfplots}
\usepackage{epstopdf}
\usepackage{float}
\usepackage{circuitikz}
\usepackage[export]{adjustbox}

\usepackage{enumitem}
\usepackage{algorithmic}
\usepackage{algorithm}

\title{Safe Reinforcement Learning for Grid Voltage Control}

\author{%
  Thanh Long Vu\thanks{Authors have equal contributions}, Sayak Mukherjee\footnotemark[1], Renke Huang, Qiuhua Huang  \\
  Pacific Northwest National Laboratory,\\
  Richland, WA, USA \\
  \texttt{\{thanhlong.vu@pnnl.gov, sayak.mukherjee@pnnl.gov,}\\ \texttt{renke.huang@pnnl.gov, qiuhua.huang@pnnl.gov\}} 
  
}

\begin{document}

\maketitle

\begin{abstract}
Under voltage load shedding has been considered as a standard approach to recover the voltage stability of the electric power grid under emergency conditions; yet this scheme usually trips a massive amount of load inefficiently. Reinforcement learning (RL) has been  adopted as a promising approach to circumvent the issues; however, RL approach usually cannot guarantee the safety of the systems under control. In this paper, we discuss a couple of novel safe RL approaches, namely constrained optimization approach and Barrier function-based approach, that can safely recover voltage under emergency events. This method is general and can be applied to other safety-critical control problems. Numerical simulations on the 39-bus IEEE benchmark are performed to demonstrate the effectiveness of the proposed safe RL emergency control. 

\end{abstract}

\section{Introduction}
\label{sec:intro}

Voltage instability is one of the major causes of power blackouts \citep{AEMO2017report}. 
Rule-based under-voltage load shedding (UVLS) is a standard emergency control to deal with voltage instability \citep{Taylor92}. Though this method is effective and fast, the lack of coordination among substations can lead to unnecessary load shedding \citep{Bai11}. Recently, reinforcement learning (RL) approaches \citep{SuttonBarto, mnih2013atari,LillicrapHPHETS15,schulman2017ppo,schulman2015trpo} has been successfully developed for emergency voltage control, while significantly reducing the amount of load shedding \citep{huang2019loadshedding_DRL, zhang2018loadshedding_DRL, huang2020accelerated}.

Though multiple RL methods were developed in the literature, the safety of the system under RL control is paid relatively less attention. In this paper, we discuss our two recently proposed distinct approaches for safe emergency voltage control, namely the constrained optimization approach \citep{vu2020safe} and Barrier function-based approach \citep{vu2021barrier}. In the constrained optimization-based safe RL method, the RL agent searches for the optimal control policy that maximizes the reward function, while obeying the safety constraint. In the Barrier function-based approach, a Barrier function is included into the reward function to guide the searching of the optimal control policy that can render the voltage to avoid safety bounds.

In a related work \citep{cheng2019endtoend}, model-based control Barrier Lyapunov function was used to design a control to compensate for the model-free reinforcement learning control in order to ensure the safety of the system. One limitation of this work lies on the limited scalability of the on-line learning of unknown system dynamics, and thus, this method was only demonstrated in simple lower-order systems. The safe RL method we present in this paper does not involve any model-based design, but only searches over the control policy space. As such, the proposed method is applicable to large scale systems. Another similar work was presented in \citep{choi2020reinforcement} where  a unified RL-based framework was used to learn the dynamic uncertainty in the control Lyapunov function, control Barrier function, and other dynamic control affine constraints altogether in a single learning process. Again, the distinction of our work is that the Barrier function was included in the reward function to guide the control learning process, without any model knowledge.

\section{Safe Reinforcement Learning Problem Formulation}
\label{sec.ARS}

We consider the RL formulation for the emergency voltage control of power systems
with a (partially observable) MDP defined by a tuple $(S,A, \mathcal{P},r,\gamma)$ \citep{SuttonBarto}. The state space (grid dynamics) $S \subseteq \mathbb{R}^n$ and action space (controllable AC motor loads) $ A \subseteq \mathbb{R}^m$ are continuous, environment transition function $\mathcal{P}: S \times A \to S$  characterizes the stochastic transition of the grid states during the dynamic events along with reward $r: S \times A \to R$, and the discount factor $\gamma \in (0,1)$. 

\begin{figure}[H]
    \centering
    \includegraphics[width = 0.7\linewidth, height = 5.1 cm]{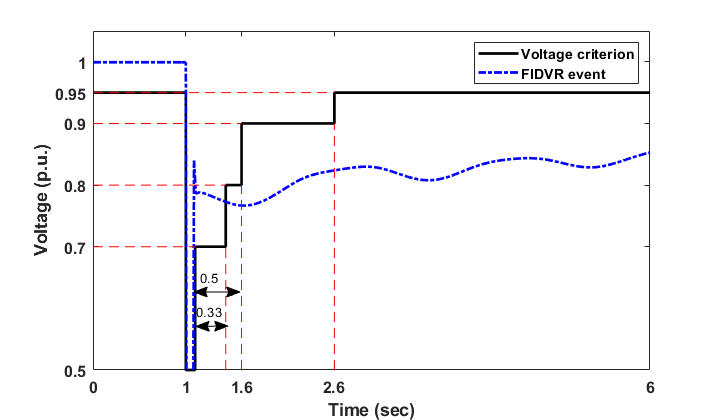}
    \caption{Safety bounds of voltages due to the transient voltage recovery criterion for transmission system}
    \label{fig.requirement}
    \vspace{-.5 cm}
    \end{figure}

{\bf Safety bounds:} In the load shedding problem, the objective is to recover the voltages of the electric power grid after the faults so that the post-fault voltages  will recover according to the standard recovery profile as shown in the Fig. \ref{fig.requirement}. 
In particular, the standard requires that, after fault clearance, voltages should return to at least $0.8, 0.9$, and $0.95$ p.u. within $0.33s, 0.5s$, and $1.5s$. The states of the power grid should obey these time-dependent bounds. Let us denote the safety set $\mathcal{C}_i \subset \mathbb{R}^n$ for the $i^{th}$ time interval $t \in T_i$ after faults, then we would require,
\begin{align}
    s_t \in \mathcal{C}_i, \;\; t \in T_i.
\end{align}

\textbf{Observation space:} Accessing all the dynamic states of the power system is a difficult task, and the operators can only measure a limited number of states and  outputs. For the voltage control problem, the bus voltage magnitudes $V(t)$ 
are easily measurable, and therefore, considered in the the observation space, along with remaining percentage of loads. Please note that with slight abuse of notations, we continue to denote partially observable states or the outputs by $s_t$.

 \textbf{Action space:} We consider controllable loads as actuators where load shedding locations are generally set by the utilities by solving a rule-based optimization problem for secure grid operation. We consider the operator can shed upto $20\%$ of the total load at a particular bus at any given time instant. The action space is continuous with $[-0.2,\;0]$ range where $-0.2$ denotes the $20\%$ load shedding.

Accordingly, the safe RL problem is to design a RL algorithm in which the RL agent learns over the action space an optimal control policy that maximizes the expected reward function, while obeying the safety bounds of power system voltages. 

We use an accelerated Augmented Random Search (ARS) algorithm \citep{huang2020accelerated} to quickly train the neural network-based control policy in a parallel computing mechanism as our algorithmic framework even before imposing any safety considerations. The ARS agent performs parameter-space exploration and estimates the gradient of the expected reward using sampled rollouts to update the control policy. In  the  load  shedding  problem,  the  objective  is  to  recover  the  voltages  of  the  electric  power  grid after  the  faults  so  that  the  post-fault  voltages  will  recover according  to  the  standard as shown in Fig. \ref{fig.requirement}. In particular, for the load shedding enabled voltage control problem, the ARS agent's objective is to maximize the expected reward, where
the reward $r_t$ at time $t$ was defined as follows:
\begin{align}
\label{eq.reward}
r(t) = 
\begin{cases}
&-1000  {\;\;\; \emph{if}\;\;\;} V_i(t)<0.95,\;\;\; T_{pf} +4<t \\
& c_1 \sum_i \Delta V_i(t) - c_2 \sum_j \Delta P_j (p.u.) - c_3 u_{ivld}, \\&{\;\;\; \emph{\mbox{otherwise}},}
\end{cases}
\end{align}
where,
\vspace{-.5 cm}
\footnotesize
\begin{align}
\Delta V_i(t) =
\begin{cases}
&\min \{V_i(t) - 0.7, 0\},  \emph{\;if\;} t \in (T_{pf}, T_{pf} +0.33) \\
&\min \{V_i(t) - 0.8, 0\}, \emph{\;if\;} t \in (T_{pf} +0.33, T_{pf} +0.5 ) \\
&\min \{V_i(t) - 0.9, 0\} , \emph{\;if\;}  t \in (T_{pf} +0.5, T_{pf} +1.5) \\
&\min \{V_i(t) - 0.95, 0\} , \emph{\;if\;} t \in t > T_{pf} +1.5.
\end{cases}\nonumber
\end{align}
\normalsize

In the reward function \eqref{eq.reward}, $T_{pf}$ is the time instant of fault clearance;
$V_i(t)$ is the voltage magnitude for bus $i$ in the power
grid;  $\Delta P_j (t)$ is the
load shedding amount in p.u. at time step $t$ for load bus $j$; 
invalid action penalty $u_{ivld}$ if the DRL agent still provides
load shedding action when the load at a specific bus has already been shed to zero at the previous time step when
the system is within normal operation. $c_1 , c_2$, and $c_3$ are the weight factors for the above three parts.

\section{Safe Reinforcement Learning Approaches}
We discuss two of our recently proposed safe RL approaches for emergency voltage control. The baseline RL algorithm used for both of the methodologies is a parallelized version of augmented random search (ARS) \citep{mania2018_ARS} developed for power system control \citep{huang2020accelerated}.  
\subsection{Constrained optimization approach}
\label{sec.safeARS}

The safety requirement of the power grid can be defined by a constraint on a time-dependent safety function $f(s_t,a_t,t)$, i.e., the system is safe if $f(s_t,a_t,t) \ge 0$. For the voltage safety requirement, the following time-dependent safety function can be used:
\begin{align}
\label{safeVol}
   f(s_t,a_t,t)= \begin{cases}
   & 0.4^2 -\max_i (V_i(t)-1.1)^2  \;\;\emph{if}\;\; T_{pf}<t<T_{pf}+0.33, \\
   & 0.35^2 -\max_i (V_i(t)-1.15)^2  \;\;\emph{if}\;\; T_{pf}+0.33<t<T_{pf}+0.5, \\
   &0.3^2 -\max_i (V_i(t)-1.2)^2  \;\;\emph{if}\;\; T_{pf}+0.5<t<T_{pf}+1.5, \\
   &0.275^2 -\max_i (V_i(t)-1.225)^2  \;\;\emph{if}\;\;  t>T_{pf}+1.5. 
   \end{cases}
\end{align}
where $V_i(t)$ is the bus voltage magnitude for bus $i$ in the power
grid. The constraint $f(s_t,a_t,t) \ge 0$ will ensure the voltage will recover as in the criterion (Fig. \ref{fig.requirement}).
The goal of a safe ARS agent is to learn the optimal control policy that maximizes the expected reward, and at the same time, satisfies the safety requirement defined by $f(s_t,a_t,t) \ge 0$. A natural approach  that was extensively used in the literature \citep{JMLR:v16:garcia15a} is to formulate this problem as the following constrained optimization  
\begin{align}
    &\max_{a_t \in A} E[ r(s_t,a_t)] \;\;
    {\bf{s.t.}} \;\;  E[f(s_t,a_t,t))] \ge 0.
\end{align}
which can be approximately solved by putting a high penalty for the safety violation \citep{KADOTA2006279}.  

In this paper, we proposed a different approach in which we consider the Lagrangian function  
\begin{align}
\label{Lfunction}
    \mathbf{L}(\pi,\lambda) = E[ r(s_t,a_t)]  + \lambda E[f(s_t,a_t,t)] 
    = E[ r(s_t,a_t) + \lambda f(s_t,a_t,t)] 
\end{align}
where $\lambda>0$ is the safety multiplier. 
Since in the original optimization we have the constraint $f(s_t,a_t,t) \ge 0$, the dual function $\mathbf{L}(\pi,\lambda)$ is an upper bound for the original optimization problem. As such, we can use the dual gradient descent method to get the optimum value of both $\mathbf{L}(\pi,\lambda)$ and the safety multiplier $\lambda$, in which we maximize $\mathbf{L}(\pi,\lambda)$ when $\lambda$ is fixed as in the standard ARS algorithm and minimize $\mathbf{L}(\pi,\lambda)$ by using gradient method to find the optimal value of $\lambda$. In other words, the objective of the safe ARS agent is to learn an optimal policy $\pi^*_{\theta}(s_t,a_t)$ from a {min-max} problem on the Lagrangian function
\begin{align}
    (\pi^*, \lambda^*) = \bf{argmax}_{a \in A} \bf{argmin}_{\lambda>0} \mathbf{L}(\pi,\lambda)
\end{align}

\subsection{Barrier Function-based Safe ARS for Emergency Load Shedding}
\label{sec.safeARS}
In the proposed method, the reward function is included with a Barrier function that will go to minus infinity when the system state tends to the safety bounds.
Accordingly, the following time-dependent Barrier function can be used:
\begin{align}
\label{safeVol}
   B(s_t,t)= \begin{cases}
   & \sum_i 1/(V_i(t)-0.7)^2) \;\;\emph{if}\;\; T_{pf}<t<T_{pf}+0.33, \\
   & \sum_i 1/(V_i(t)-0.8)^2) \;\;\emph{if}\;\; T_{pf}+0.33<t<T_{pf}+0.5, \\
   & \sum_i 1/(V_i(t)-0.9)^2)  \;\;\emph{if}\;\; T_{pf}+0.5<t<T_{pf}+1.5, \\
   & \sum_i 1/(V_i(t)-0.95)^2)  \;\;\emph{if}\;\; t>T_{pf}+1.5. 
   \end{cases}
\end{align}
where $V_i(t)$ is the bus voltage magnitude for bus $i$ in the power
Now, the reward function we consider in the safe ARS algorithm is as follows:
\begin{align}
\label{reward}
    R(t) = r(t) - c_4 B(s_t,t),
\end{align}
where $r(t)$ is the reward, $B(s_t,t)$ is defined in \eqref{safeVol}, and $c_4>0$ is a weight factor. \\

{\bf{Safety Considerations:}} In this safe ARS algorithm, as the ARS agent learns the optimal control policy that maximizes the reward function, in each iteration it will search over the control policy space and select, among several directions, the best-performance policies where the reward function is highest. Also, when going to the next iteration, the control policy is updated by the surrogate gradient-like method on the reward function. As such, the reward function is increasing in expectation during the update of control policy. Hence, during the exploration and update of the control policy, the reward function cannot tend to minus infinity. As a result, the Barrier function cannot tend to minus infinity during the exploration and update of the control policy. Therefore, the safety bounds of system state are not violated during the searching process of the optimal control policy. 

Mathematically, in the ARS-based learning process for the optimum control policy, the expected reward is lower bounded. This means that, along the trajectory of the agent state, there is only a zero-measure set of samples in which the reward function can go to infinity. Hence, the reward function is bounded almost surely, i.e., $\mathbf{P}\{R_t >-\infty, \forall t\}=1$. Hence, the Barrier function is also bounded almost surely along the trajectory $s(t).$ 
We note that when $s(t)$ goes through the safety bounds, then the Barrier function will go to minus infinity. Therefore, we can conclude that the voltages will not violate the safety bounds almost surely. 

\subsection{Algorithmic Description of Safe RL for Voltage Control}
\label{sec:intro}
Algorithm $1$ describes the constrained optimization based approach, and algorithm $2$ presents the optimal control design by the Barrier function-based safe ARS method described hereby. 
\begin{algorithm}[]
\caption{Constrained Optimization based Safe ARS:}
\begin{algorithmic}
\STATE 1. \textbf{Hyperparameters:} Step size $\alpha$, number of policy perturbation directions per iteration $N$, standard deviation of the exploration noise $\nu$, number of top-performing perturbed directions selected for updating weights $b$, number of rollouts per perturbation direction $m$. Decay rate $\epsilon$. 
\STATE 2. \textbf{Initialize:} Policy weights $\theta_0$ with small random numbers; initial safety multiplier in the reward function $\lambda_0,$ the running mean of observation states $\mu_0 = 0 \in R^n$ and the running standard deviation of observation states $\Sigma_0 = I_n \in R^n$, where $n$ is the dimension of observation states, the total iteration number $H$.
\FOR{ iteration $t \leq H$}
\STATE 3. Sample $N$ number of random directions $\delta_{1},\dots,\delta_{N} \in \mathbb{R}^{n_{\theta}}$ with the same
dimension as policy weights $\theta$.
\FOR{each $\delta_{i}, i=1,\dots, N$}
\STATE 4. Add perturbations to policy weights: $\theta_{ti+} = \theta_{t - 1} +
\nu \delta_i$ and $\theta_{ti-} = \theta_{t - 1} - \nu \delta_i$
\STATE 5. Do total $2m$ rollouts (episodes) denoted by $R_{p\in T} (.)$
for different tasks $p$ sampled from task set $T$ corresponding to $m$ different faults with the $\pm$ perturbed policy weights. Calculate safety functions for all $\pm$ perturbations and calculate the average rewards of $m$ rollouts as the rewards for $\pm$ perturbations, i.e., $\bar{R}_{ti +}$ and $\bar{R}_{ti -}$ are 
\begin{align}
    &\bar{R}_{ti+} = \frac{1}{m} R_{p\in T}(\theta_{ti+},\mu_{t - 1},\Sigma_{t - 1}),\\
    &\bar{R}_{ti-} = \frac{1}{m} R_{p\in T}(\theta_{ti-},\mu_{t - 1},\Sigma_{t - 1})
\end{align}
where the reward function $R$ is defined as the combined reward function in \eqref{Lfunction}, i.e., $R(.) = r(s_t,a_t) + \lambda_t f(s_t,a_t,t).$
\STATE 6. During each rollout, states $s_{t,k}$ at time step $k$ are first normalized and then used as the input for inference with policy $\pi_{\theta_t}$ to obtain the action $a_{t,k}$, which is applied to the environment and new states $s_{t,k+1}$ is returned, as shown in (3). The running mean $\mu_t$ and standard deviation $\Sigma_t$ are updated with $s_{t,k+1}$
\begin{align}
    &s_{t,k} = \frac{s_{t,k} - \mu_{t-1}}{\Sigma_{t-1}},\\
    & a_{t,k} = \pi_{\theta_t}(s_{t,k}),\\
    & s_{t,k+1} \gets \mathcal{P}(s_{t,k},a_{t,k}) ,
\end{align}
\ENDFOR
\STATE 7. Sort the directions based on $\max ( \bar{R}_{ti + } , \bar{R}_{ti - } )$, select
top $b$ directions, calculate their standard deviation $\sigma_b$
\STATE 8. Update the policy weight:
\begin{align}
    \theta_{t+1} = \theta_{t} + \frac{\alpha}{b  \sigma_{b}} \sum_{i=1}^{b} (\bar{R}_{ti+} - \bar{R}_{ti-})\delta_{i}
\end{align}
\STATE 9. Step size $\alpha$ and standard deviation of the exploration noise $\nu$ decay with rate $\epsilon$: $\alpha \gets \epsilon \alpha, \nu\gets \epsilon \nu$.
\STATE 10. Update of the safety multiplier: Check the safety violation for all the rollouts. If there is any safety violation: $\lambda_{t+1} \gets 2 \lambda_t$. Otherwise: $\lambda_{t+1} \gets \lambda_t /2.$
\ENDFOR
\RETURN 11.  Return $\theta$,  and {$\lambda$}.
\end{algorithmic}
\end{algorithm}

\begin{algorithm}[t!]
\caption{Barrier function-based safe ARS:}
\begin{algorithmic}
\STATE 1. \textbf{Hyperparameters:} Step size $\alpha$, number of policy perturbation directions per iteration $N$, standard deviation of the exploration noise $\nu$, number of top-performing perturbed directions selected for updating weights $b$, number of rollouts per perturbation direction $m$. Decay rate $\epsilon$. 
\STATE 2. \textbf{Initialize:} Policy weights $\theta_0$ with small random numbers; initialize the running mean of observation states $\mu_0 = 0 \in R^n$ and the running standard deviation of observation states $\Sigma_0 = I_n \in R^n$, where $n$ is the dimension of observation states, the total iteration number $H$.
\FOR{ iteration $t \leq H$}
\STATE 3. Sample $N$ number of random directions $\delta_{1},\dots,\delta_{N} \in \mathbb{R}^{n_{\theta}}$ with the same
dimension as policy weights $\theta$.
\FOR{each $\delta_{i}, i=1,\dots, N$}
\STATE 4. Add perturbations to policy weights: $\theta_{ti+} = \theta_{t - 1} +
\nu \delta_i$ and $\theta_{ti-} = \theta_{t - 1} - \nu \delta_i$
\STATE 5. Do total $2m$ rollouts (episodes) denoted by $R_{p\in T} (.)$
for different tasks $p$ sampled from task set $T$ corresponding to $m$ different faults with the $\pm$ perturbed policy weights. Calculate the average rewards of $m$ rollouts as the rewards for $\pm$ perturbations, i.e., $\bar{R}_{ti +}$ and $\bar{R}_{ti -}$ are 
\begin{align}
    &\bar{R}_{ti+} = \frac{1}{m} R_{p\in T}(\theta_{ti+},\mu_{t - 1},\Sigma_{t - 1}),\\
    &\bar{R}_{ti-} = \frac{1}{m} R_{p\in T}(\theta_{ti-},\mu_{t - 1},\Sigma_{t - 1})
\end{align}
where the reward function $R$ is defined as the combined reward function in \eqref{reward}.

\STATE 6. During each rollout, states $s_{t,k}$ at time step $k$ are first normalized and then used as the input for inference with policy $\pi_{\theta_t}$ to obtain the action $a_{t,k}$, which is applied to the environment and new states $s_{t,k+1}$ is returned, as shown in (3). The running mean $\mu_t$ and standard deviation $\Sigma_t$ are updated with $s_{t,k+1}$
\begin{align}
    &s_{t,k} = \frac{s_{t,k} - \mu_{t-1}}{\Sigma_{t-1}},\\
    & a_{t,k} = \pi_{\theta_t}(s_{t,k}),\\
    & s_{t,k+1} \gets \mathcal{P}(s_{t,k},a_{t,k}) ,
\end{align}
\ENDFOR
\STATE 7. Sort the directions based on $\max ( \bar{R}_{ti + } , \bar{R}_{ti - } )$, select
top $b$ directions, calculate their standard deviation $\sigma_b$
\STATE 8. Update the policy weight:
\begin{align}
    \theta_{t+1} = \theta_{t} + \frac{\alpha}{b  \sigma_{b}} \sum_{i=1}^{b} (\bar{R}_{ti+} - \bar{R}_{ti-})\delta_{i}
\end{align}
\STATE 9. Step size $\alpha$ and standard deviation of the exploration noise $\nu$ decay with rate $\epsilon$: $\alpha \gets \epsilon \alpha, \nu\gets \epsilon \nu$.
\ENDFOR
\RETURN 10.  Return $\theta$.
\end{algorithmic}
\end{algorithm}

\begin{itemize}
    \item The safe ARS learner is an actor at the top to delegate tasks and collect returned information, and updates policy weights $\theta$. For the algorithm $1$, it also controls the update of the safety multiplier $\lambda$.
    \item The learner communicates with subordinate workers and each of these workers is responsible for one or more perturbations (random search) of the policy weights as in Step 4.
    \item In Step 7, the ARS learner combines the results from each worker calculated in Step 5 (which include the average reward of multiple rollouts), sorts the directions according to the reward, selects the  best-performing directions.
    \item Then, in Step 8, for both of the algorithms, the ARS learner updates the policy weights centrally based on the perturbation results from the top performing workers. For Algorithm $1$, in Step 10, the safety multiplier $\lambda$ is centrally updated based on the perturbation results, and checking the safety violations. For simplicity, we proposed a more heuristic approach to find the sub-optimal value of the safety multiplier $\lambda$, in which we will check the safety conditions in each iteration. If the safety constraint is not violated in the iteration, we reduce the value of $\lambda$ two times. Otherwise, we increase the value of $\lambda$ two times. 
    \item The workers do not execute environment rollout tasks by themselves. They spawn a number of actors and assign these tasks to these subordinate  actors. Note that each worker needs to collect the rollout results from multiple tasks inferring with the same perturbed policy, and each actor is only responsible for one environment rollout with the specified task and perturbed policy sent by its up-level worker. For the environment rollouts, power system dynamic simulations are performed in parallel. 
\end{itemize}

\section{Test Results}
\label{sec.results}
\begin{wrapfigure}{r}{0.5\textwidth}
 \centering
    \includegraphics[width=0.48\textwidth]{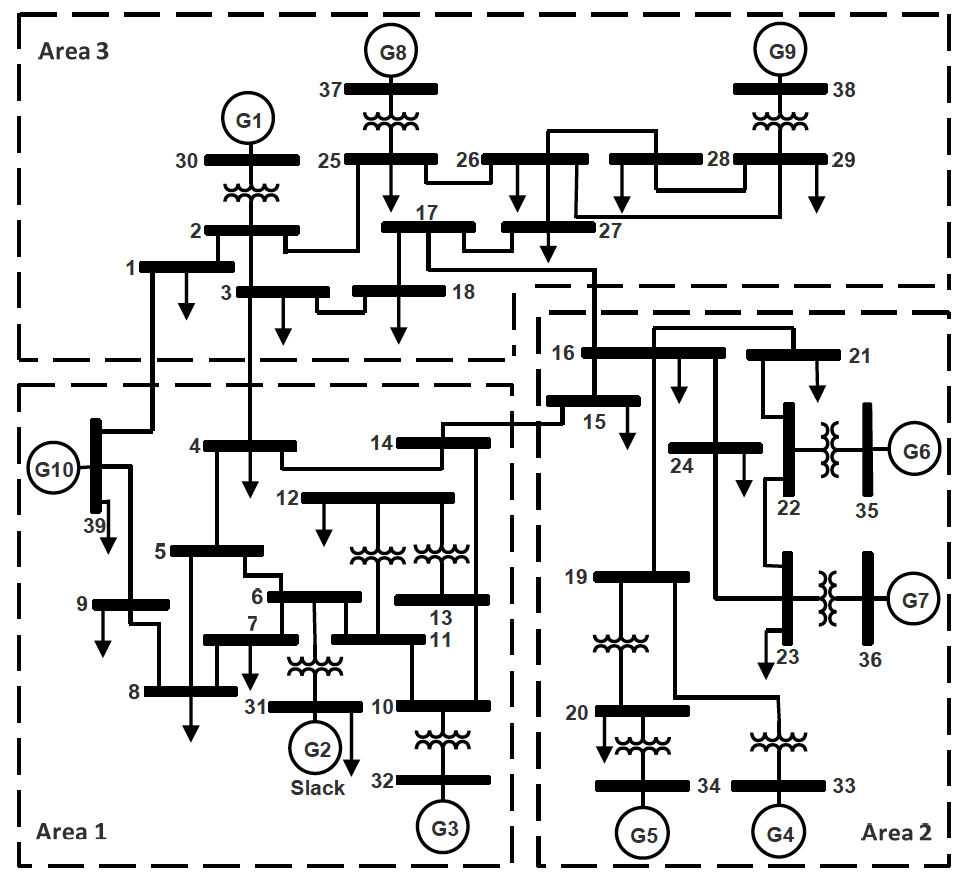}
    \caption{IEEE 39-bus system}
    \label{fig:39bus}
\end{wrapfigure}

We perform simulations in the IEEE benchmark $39-$bus, $10-$generator model as in Fig. \ref{fig:39bus}. The simulations performed in a Linux mainframe with $27$ cores. The power system simulator runs using GridPack\footnote{https://www.gridpack.org}, 
and the safe deep RL algorithm is implemented in a separate platform using Python. A software setup has been built such that the grid simulations in the GridPack and the RL iterations in the python can communicate.  

\begin{figure}
\begin{minipage}[b]{.5\textwidth}
  \centering
  \includegraphics[width=.8\linewidth]{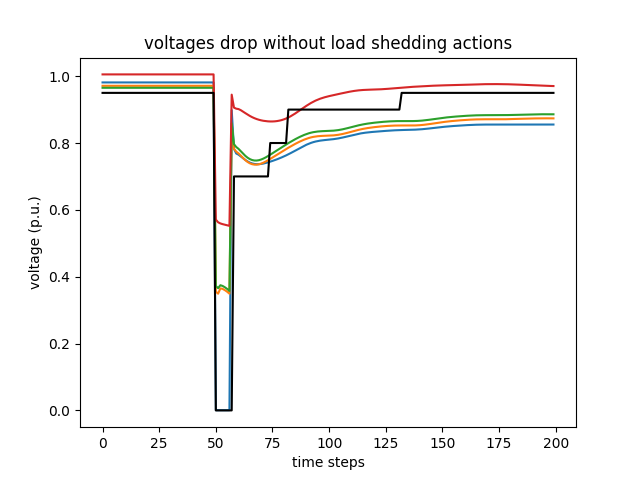}  
  \caption{Voltage drops after fault happens at bus 4 without any emergency control measures}
  \label{fig:nocontrol}
  \end{minipage}
\begin{minipage}[b]{.5\textwidth}
  \centering
  \includegraphics[width=.8\linewidth]{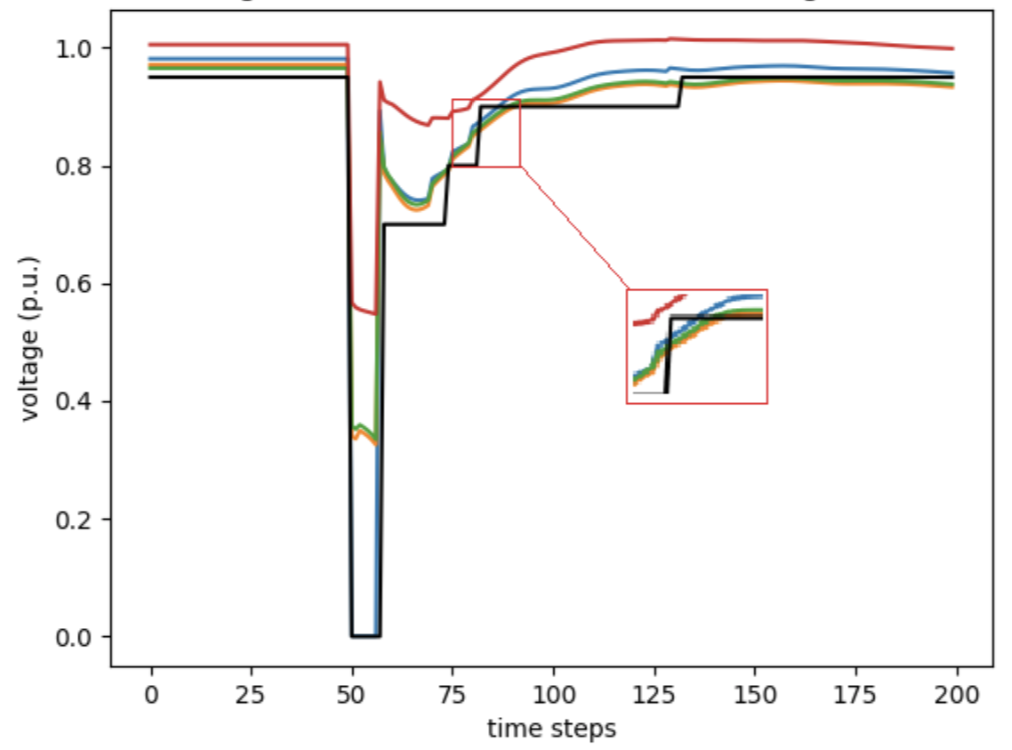}  
  \caption{Voltage profile under the standard ARS-based emergency load shedding}
  \label{fig:standardARS}
  \end{minipage}


\begin{minipage}[b]{.5\textwidth}
  \centering
  \includegraphics[width=.8\linewidth]{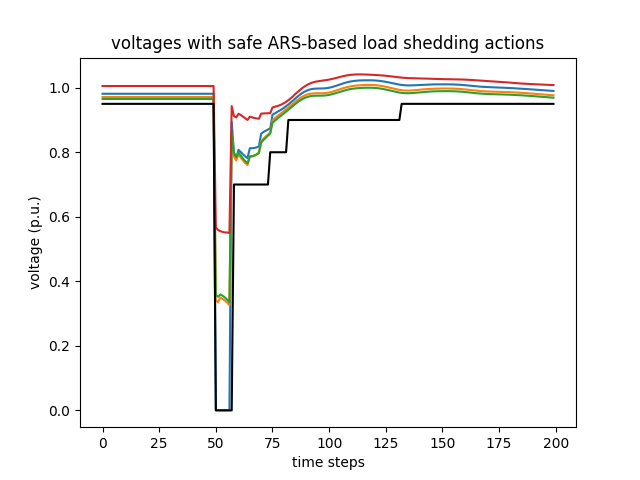}  
  \caption{Voltage profile under the constrained optimization emergency load shedding response}
  \label{fig:constrainedoptimization}
  \end{minipage}
\begin{minipage}[b]{.5\textwidth}
  \centering
  \includegraphics[width=.8\linewidth]{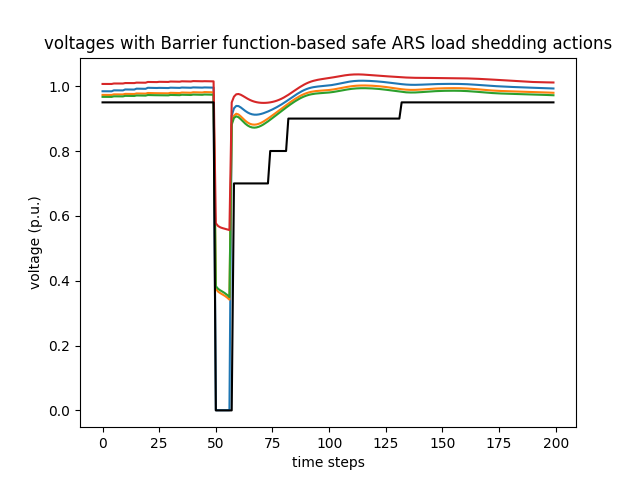}  
  \caption{Voltage profile under the Barrier function-based emergency load shedding}
  \label{fig:Barrier}
  \end{minipage}
\vspace{-1 cm}
 \end{figure}

\begin{figure}[H]
\begin{minipage}[b]{.5\textwidth}
  \centering
  \includegraphics[width=.8\linewidth]{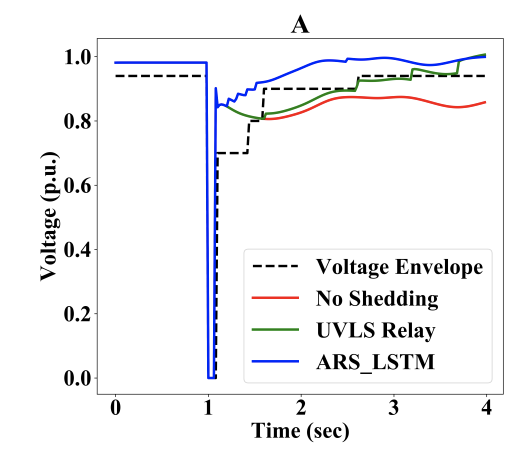}  
  \caption{Effectiveness of the ARS-based RL scheme}
  \label{fig:comparison}
  \end{minipage}
\begin{minipage}[b]{.5\textwidth}
  \centering
  \includegraphics[width=.8\linewidth]{volt_AI_FNN_fullscale_faultbus4_edited.png}  
  \caption{Voltage profile under the standard ARS-based emergency load shedding after fault happens at bus 7}
  \label{fig:ARS1}
  \end{minipage}


\begin{minipage}[b]{.5\textwidth}
  \centering
  \includegraphics[width=.8\linewidth]{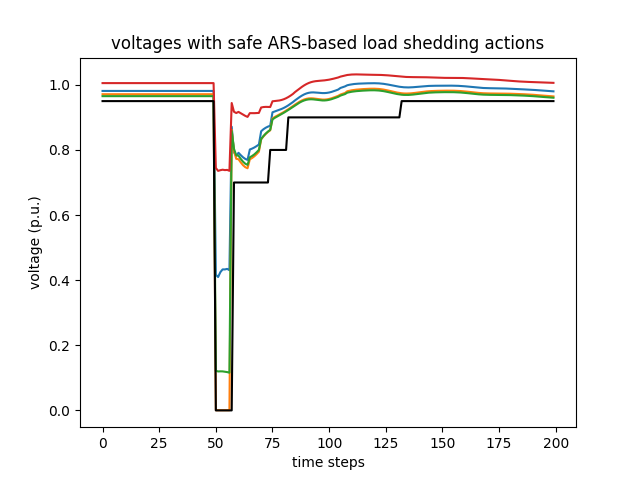}  
  \caption{Voltage profile under the constrained optimization emergency load shedding response after fault happens at bus 7}
  \label{fig:ARS2}
\vspace{4ex}
  \end{minipage}
\begin{minipage}[b]{.5\textwidth}
  \centering
  \includegraphics[width=.8\linewidth]{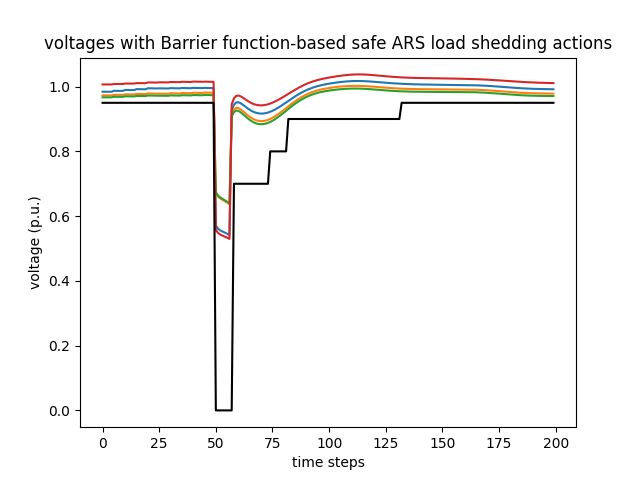}  
  \caption{Voltage profile under the Barrier function-based emergency load shedding after fault happens at bus 7}
  \label{fig:safeARS2}
  \vspace{4ex}
  \end{minipage}
\vspace{-1 cm}
 \end{figure}

The policy generates required optimized load shedding actions at buses $4, 7,$ and $18$ to meet the voltage recovery requirements. 
Observations included voltage magnitudes at buses $4, 7, 8,$ and $18$ as well as the remaining fractions of loads served by buses $4, 7$ and $18$. The control action for buses $4, 7,$ and $18$ at each action time step was a continues variable from $0$ (no load shedding) to $-0.2$ (shedding $20\%$ of the initial total load at the bus). We fix a the task set T consisting of nine different tasks (fault scenarios) for the training purposes. The fault scenarios began with the flat start in the dynamic simulation. At $1.0$ s, we apply short circuit faults at one of the bus $4, 15$, or $21$ with a fault duration of $0.0$ s (no fault), $0.15$ s, or $0.28$ s and the fault was self-cleared. 
When there is no additional feedback control implemented in the system, we can observe voltage recovery profile largely violates the required recovery bounds as in Fig. \ref{fig:nocontrol}. Fig. \ref{fig:comparison} shows the performance comparison between standard ARS-based RL scheme 
and under-voltage load shedding (UVLS ) for a test task with 0.08 s of fault at bus 3, where the voltage with UVLS control (green
curve) at bus 4 could not recover while
standard-ARS based control (blue curve) could recover to meet the standard. Although standard ARS based RL outperforms the traditional UVLS schemes, it may still encounter safety issues due to the lack of dedicated safety measures. The performances of the standard ARS-based load shedding and safe RL-based emergency control are depicted in Figs. \ref{fig:standardARS}- \ref{fig:Barrier} for fault at bus $4$ with $0.15$s duration.
These figures show that the safe RL-based load shedding can perform much better than the standard ARS-based load shedding in meeting the safety requirement described by the transient voltage recovery criterion depicted in Fig. 1. In particular, without any Safety consideration based RL design, the voltages at buses $7,8,$ and $18$ could not recover to $0.9$ p.u. within $0.5$s after the fault clearance, and the voltages at buses $8$ and $18$ do not recover to $0.95$ p.u. within $1.5$s after the fault-cleared time. Yet, the safe ARS-based load shedding using both the methods makes the voltages at all buses $4,7,8,18$ recover well as required. 
 To test the adaptation of the safe ARS, we consider a fault not encountered during the training, in which short-circuit fault happens at bus $7$ at $1.0$s with a fault duration of $0.15$s and then the fault self-cleared. The performances of standard ARS-based load shedding and safe ARS-based load shedding using constrained optimization and Barrier-function based approaches are depicted in Figs. \ref{fig:ARS1} - \ref{fig:safeARS2} substantiating our designs.
\section{Conclusions}

In this paper, we discussed two safe deep reinforcement learning approaches for power system voltage stability control using load shedding. Remarkably, by incorporating a constrained optimization and a Barrier function into the reward function, the safe ARS algorithm resulted in a control policy that could prevent the system state from violating the safety bounds, and hence, enhance the safety of the electric power grid during the load shedding. Case studies on the IEEE $39$-bus demonstrated that safe ARS-based load shedding scheme successfully recovers the voltage stability of power systems even in events it did not see during the training. In addition, in comparison to the standard ARS-based load shedding, it showed advantages in both safety level and fault adaptability.

\begin{ack}
Pacific Northwest National Laboratory (PNNL) is operated by Battelle for the U.S. Department of Energy (DOE) under Contract DE-AC05-76RL01830. This work was funded by DOE ARPA-E OPEN 2018 Program.
\end{ack}

\bibliography{refs}
\bibliographystyle{abbrvnat}

\newpage
\appendix

\end{document}